%% file: root.tex
\documentclass[letterpaper, 10 pt, conference]{ieeeconf}  

\IEEEoverridecommandlockouts                                                  
\title{\LARGE \bf
Human2Humanoid: Physics-Aware Cross-Morphology Motion Retargeting for Humanoid Robots
}

\author{Tianchen Huang, Feiyang Yuan, Junchi Gu, Shurui Fang, Xiaohu Zhang, Yu Wang, Wei Gao and Shiwu Zhang
\thanks{This work was supported in part by the National Science and Technology Major Project (Grant No. 2026ZD1609100), in part by the National Natural Science Foundation of China under Grants U22B2040, in part by the Key Science \& Technology Project of Anhui Province (202523j08050001), in part by the Major Project of Anhui Province's Science and Technology Innovation Breakthrough Plan (202423h08050003), and in part by the Fundamental Research Funds for the Central Universities with grant No. YD2090002019. (Corresponding authors: Wei Gao; Yu Wang.)}%
\thanks{The authors are with the Institute of Humanoid Robots, Department of Precision Machinery and Precision Instrumentation, University of Science and Technology of China, Hefei, Anhui 230026, China. {\tt\footnotesize weigao@ustc.edu.cn; wangyuustc@ustc.edu.cn}}%
}

\usepackage{siunitx}
\usepackage{bbm}
\usepackage{etoolbox}
\usepackage{multicol}
\usepackage{cite}
\usepackage[bookmarks=true]{hyperref}
\usepackage{graphics} % for pdf, bitmapped graphics files
\usepackage{times} % assumes a new font selection scheme installed
\usepackage{amsmath} % assumes amsmath package installed
\usepackage{amssymb}  % assumes amsmath package installed
\usepackage{svg}
\usepackage{mathrsfs}
\usepackage{amsfonts}
\usepackage{wrapfig}
\usepackage{amsmath}

\usepackage{kantlipsum}
\usepackage{subcaption}
\usepackage{marvosym}

\usepackage{graphicx}
\usepackage{float}
\usepackage{ctable}
\usepackage{cuted}
\usepackage{colortbl}
\usepackage{multirow}
\usepackage{wrapfig}
\usepackage[misc,geometry]{ifsym}
\usepackage{algorithm}
\usepackage{algpseudocode}
\usepackage{xcolor}   % For color
\usepackage{pifont}    % For check and cross symbols

\usepackage{bm}
\usepackage{caption}
\usepackage{setspace}

\usepackage{enumitem}
\usepackage{threeparttable}
\newcommand{\TODO}[1][]{\textcolor{red}{\bf [TODO]}}
\setlist[enumerate,1]{itemsep=3pt}

\definecolor{formalgreen}{rgb}{0.1, 0.7, 0.1}  % A darker, more formal green
\definecolor{formalred}{rgb}{0.9, 0.2, 0.2}  % A darker, more formal green

\begin{document}

\maketitle
\thispagestyle{empty}
\pagestyle{empty}

%%%%%%%%%%%%%%%%%%%%%%%%%%%%%%%%%%%%%%%%%%%%%%%%%%%%%%%%%%%%%%%%%%%%%%%%%%%%%%%%
\begin{abstract}
\input{tex/0-abs}
\end{abstract}
%%%%%%%%%%%%%%%%%%%%%%%%%%%%%%%%%%%%%%%%%%%%%%%%%%%%%%%%%%%%%%%%%%%%%%%%%%%%%%%%
\section{Introduction}
\input{tex/1-intro}
%%%%%%%%%%%%%%%%%%%%%%%%%%%%%%%%%%%%%%%%%%%%%%%%%%%%%%%%%%%%%%%%%%%%%%%%%%%%%%%%
\section{Related Work}
\input{tex/2-related_work}
%%%%%%%%%%%%%%%%%%%%%%%%%%%%%%%%%%%%%%%%%%%%%%%%%%%%%%%%%%%%%%%%%%%%%%%%%%%%%%%%
\section{Method\label{sec:method}}
\input{tex/3-method}
\section{Experiments}
\input{tex/4-exp}
\section{Conclusion}
\input{tex/5-conclusion}
%%%%%%%%%%%%%%%%%%%%%%%%%%%%%%%%%%%%%%%%%%%%%%%%%%%%%%%%%%%%%%%%%%%%%%%%%%%%%%%%
% {\small
\bibliographystyle{IEEEtran}
\bibliography{references}
% }
%%%%%%%%%%%%%%%%%%%%%%%%%%%%%%%%%%%%%%%%%%%%%%%%%%%%%%%%%%%%%%%%%%%%%%%%%%%%%%%%
% \addtolength{\textheight}{-10cm}   % This command serves to balance the column lengths
% \clearpage
% \appendix
% \input{tex/6-supp}

\end{document}

%% file: tex/0-abs.tex
Retargeting human motion to humanoid robots is critical for teleoperation, imitation learning and human-robot interaction. However, it remains challenging because of substantial morphological discrepancies between humans and robots, including differences in skeletal topology, limb proportions and degrees of freedom, as well as the scarcity of paired motion data. This paper presents Human2Humanoid, an unsupervised motion retargeting framework that transfers human motions to humanoid robot behaviors with high fidelity. To bridge the domain gap under unpaired data, we adopt a CycleGAN-based architecture equipped with a skeleton-aware graph convolutional network to capture topology-dependent motion features. To address cross-domain scale mismatches, we introduce a morphology-invariant end-effector consistency loss that aligns normalized end-effector trajectories to preserve motion semantics across embodiments. To improve physical plausibility and reduce contact artifacts, we impose explicit physics-aware feasibility constraints to encourage reproduction of the contact patterns in the source motion. Experimental results show that the proposed method successfully retargets human motion to the Unitree G1 humanoid robot without paired data, and outperforms existing methods in both downstream controllability and physical feasibility.

%% file: tex/1-intro.tex
The rapid development of humanoid robots has opened new frontiers in teleoperation, imitation learning and human-robot interaction~\cite{deepmimic, chen2025gmt, Human–RobotInteraction}. A fundamental requirement of these applications is the ability to transfer rich human motion priors into executable robot behaviors, which is also known as motion retargeting. In modern learning-based pipelines, motion retargeting 
% from humans to humanoid robots 
has served as the critical bridge between human demonstration and robot control. However, it remains highly challenging because of substantial embodiment gaps arising from two main aspects: morphological discrepancies in
skeletal topology, limb proportions and degrees of freedom (DoFs)~\cite{1998Retargetting, 2020Skeleton-aware}, and the scarcity of high-quality paired data, namely synchronized human-robot motion sequences~\cite{choi2021selfsupervis}. 

Traditional retargeting methods rely on inverse kinematics and constrained optimization to enforce robot-specific feasibility constraints, such as joint limits, body balance, collision avoidance and contact conditions~\cite{ayusawa2017motion, darvish2019}. 
Although effective and interpretable, these methods are sensitive to initial conditions, objective weights and embodiment mismatches, and may produce semantically distorted or contact-inconsistent robot motions~\cite{Globalinversekinematics, gmr}.
Data-driven methods offer an alternative by learning human-robot motion mappings, but supervised approaches depend heavily on paired human-robot motion data, which is difficult to acquire and scale across different embodiments~\cite{zhang2023, HuLei2024Pose}. 
On the other hand, unpaired retargeting methods based on adversarial or cycle-consistency objectives have been explored in computer graphics and animation~\cite{2017Unpaired, villegas2018neural, zhao2023posetomotion}. 
Nevertheless, direct application of these methods to humanoid robots is unrealistic because they mainly focus on visual plausibility but not executable joint-space trajectories, stable ground contacts or robot-specific kinematic feasibility.

%Data-driven methods offer appealing alternatives by learning the mapping from human motion to robot motion directly. Supervised approaches~\cite{zhang2023, HuLei2024Pose} treat retargeting as a regression problem from human pose representations to robot joint configurations, thereby reducing hand-crafted engineering. Yet their success depends heavily on large-scale paired datasets, which are difficult to acquire and scale across robot embodiments. To reduce this dependence, unsupervised domain adaptation methods based on CycleGAN have been explored in motion retargeting~\cite{2017Unpaired, zhao2023posetomotion, villegas2018neural, yan2024imitation}. These methods learn cross-domain mappings through adversarial and cycle-consistency objectives without requiring explicit pairing. Despite their success in computer graphics, their direct application to humanoid robots remains limited. Many existing frameworks represent the skeleton as a simple vector or image, which weakens the model's ability to exploit topological joint relationships~\cite{2020Skeleton-aware}. In addition, many animation-oriented methods are evaluated primarily by visual plausibility and often output Cartesian skeletal motion for rendered characters, whereas humanoid robots require executable joint-angle trajectories under embodiment-specific kinematic and contact constraints.

Therefore, motion retargeting through learning-based methods for humanoid robots must address two coupled challenges. The first is semantic preservation under severe morphology mismatches. When body sizes, limb proportions and joint structures differ substantially across embodiments, pose-level geometric alignment alone is insufficient, because the same action semantics may correspond to very different absolute coordinates and reachable configurations. The second is physical plausibility under unpaired learning. Distribution-level alignment alone cannot prevent contact-related artifacts such as foot skating and ground penetration, which may compromise executability on real robots. Furthermore, these two challenges are tightly coupled: methods that focus only on feasibility may distort the semantics of source motion, while methods that focus only on appearance or distribution matching may produce physically unreliable robot behaviors.

To address these challenges, this paper proposes \textbf{Human2Humanoid}, an unpaired motion retargeting framework for humanoid robots. The central idea is to cast retargeting as unpaired cross-domain transfer anchored by morphology-invariant semantic structure and explicit physical priors. Specifically, it builds upon a CycleGAN-style architecture~\cite{2017Unpaired} and incorporates a Skeleton-Aware Graph Convolutional Network to model the native topological structure of skeletal motions. To preserve action semantics across embodiments with large scale mismatches, a \textit{Morphology-Invariant End-Effector Consistency Loss} is introduced to align end-effector trajectories that are normalized relative to the T-pose of each embodiment rather than directly matching absolute coordinates. To improve physical plausibility during generation, \textit{Physics-Aware Feasibility Constraints} are further imposed to reproduce the contact patterns of source motion while suppressing foot skating and ground penetration. Consequently, the proposed framework learns cross-embodiment motion transfer without relying on paired human-robot data, and becomes more compatible with downstream control tasks.

% The proposed human-to-humanoid motion retargeting method 
The proposed method is validated on the Unitree G1 platform, with the experimental results showing that Human2Humanoid can outperform representative baselines in both semantic fidelity and physical realism. In addition, the ablation study confirms that the physics-aware training objectives make essential contributions to the overall performance.
The main contributions of this work are summarized as:
\begin{itemize}
    \item Proposing \textit{Human2Humanoid}, an unpaired motion retargeting framework for heterogeneous humanoid robots that reduces dependence on robot-specific paired datasets.
    \item Introducing a \textit{Morphology-Invariant End-Effector Consistency Loss} that preserves motion semantics under large cross-embodiment scale mismatches by aligning end-effector trajectories that are normalized relative to each embodiment's T-pose.
    \item Incorporating \textit{Physics-Aware Feasibility Constraints} into the generative training process, improving physical plausibility by mitigating contact-related artifacts such as foot skating, body floating and ground penetration.
\end{itemize}

%% file: tex/2-related_work.tex
\subsection{Motion Retargeting in Robotics}

Motion retargeting for humanoid robots has traditionally been addressed through inverse kinematics and constrained optimization. 
% These methods explicitly encode robot-specific feasibility requirements, such as joint limits, body balance, collision avoidance and contact constraints. 
Ayusawa and Yoshida jointly optimize morphology parameters and robot motion to reduce geometric mismatches while reproducing source-motion characteristics~\cite{2017Motion}. 
Penco \emph{et al.} formulate real-time whole-body retargeting as a constrained inverse kinematics and quadratic programming problem for humanoid teleoperation~\cite{2018Robust}. 
More recently, Araujo \emph{et al.} show that retargeting artifacts such as foot skating, self-collision and physical infeasibility can substantially degrade downstream humanoid motion tracking, and propose General Motion Retargeting (GMR) to combine non-uniform local scaling with constrained optimization~\cite{gmr}. 
These optimization-based methods are interpretable and can impose physical constraints directly. 
However, they remain sensitive to initial conditions, objective weights and per-motion tuning. 
Under large morphology mismatches, geometric fitting can produce poses that are reachable but semantically distorted, such as compressed strides, shifted arm trajectories or unstable stance phases.

Learning-based methods aim to reduce this engineering burden by learning human-robot mappings from data. 
A central difficulty, however, is that supervised training requires paired human-robot motion data, which are expensive to collect and difficult to scale across different embodiments. 
S3LE reduces manual data collection through a self-supervised generation procedure, but its learning process still depends on paired human poses and robot configurations~\cite{choi2021selfsupervis}. 
Moreover, its evaluation focuses mainly on upper-body motion of the COMAN humanoid robot, leaving stable lower-body control such as root displacement and foot-ground contacts not addressed. 
Recent neural retargeting methods further improve feasibility by constructing physically refined supervision. 
For example, NMR proposes a Clustered-Expert Physics Refinement pipeline that curates human motions, performs optimization-based retargeting and filtering, and uses expert-policy rollouts in simulation to generate physics-consistent human-robot motion pairs~\cite{nmr}. 
This strategy improves tracking quality, but its effectiveness still depends on morphology-specific paired supervision generated within a pre-curated robot-feasible motion subspace. 
% Therefore, extending it to a new embodiment requires rebuilding a data construction pipeline involving motion filtering, retargeting, expert-policy training and simulator-based refinement. 
Therefore, extending it to a new robot embodiment requires rebuilding the same robot-specific data-construction pipeline, with its key stages reconfigured for the target morphology and dynamics.
Moreover, because the source motions are filtered to retain the ones that the robot can in principle execute and the paired targets are further repaired for tracking feasibility, the learned mapping may be biased toward dynamically trackable motions rather than preserving the full diversity and fine-grained semantics of open-ended human motions. 
This limitation is especially relevant for motions involving explicit environmental geometry or external contacts, such as sitting, stepping onto stairs or object manipulation. 

Generative models provide another direction for cross-embodiment retargeting. 
G-DReaM represents heterogeneous embodiments with graph structures and uses energy-guided retargeting losses to train a graph-conditioned diffusion model when ground-truth target motions are unavailable~\cite{G-DReaM}. 
This direction improves scalability across embodiments, but reliable humanoid robot deployment still requires semantic structure preservation, contact consistency and kinematic feasibility in target motions. 

Overall, existing methods for robots either rely on explicit optimization, which is interpretable but sensitive to tuning, or learned mappings, which often reintroduce paired supervision through data collection, curation and physics-based refinement. This motivates an unpaired retargeting framework that eliminates the need for paired supervision while preserving cross-morphology semantics and enforcing contact-related physical plausibility.
%This motivates an unpaired retargeting framework that can reduce dependence on paired motion data while explicitly preserving cross-morphology semantics and enforcing contact-related physical plausibility.

\subsection{Motion Retargeting in Animation and Computer Graphics}

Motion retargeting has also been extensively studied in computer graphics and animation with the goal of reusing motion assets across characters with different body shapes or skeletons. 
Early works formulate the problem as constrained motion editing. 
For example, Gleicher uses spacetime constraints to preserve key motion properties such as foot-ground contacts while adapting motions across characters~\cite{1998Retargetting}. 
Since these formulations are closely related to the optimization methods used in robotic retargeting, they share similar shortcomings. 
% we focus here on later animation-oriented methods that reduce the dependence 
% and depend heavily on paired source-target motion supervision.

Learning-based retargeting in animation provides an important reference for unpaired motion transfer. 
Neural Kinematic Networks combine an analytic forward-kinematics layer with cycle-consistency and adversarial objectives, enabling unsupervised retargeting without paired motion sequences~\cite{villegas2018neural}. 
PMnet separates local pose transfer from global movement adaptation, reducing drift and distortion caused by aligning motions only in joint space~\cite{2019PMnet}. 
Skeleton-Aware Networks further introduce topology-aware convolution, pooling and unpooling operators to encode homeomorphic skeletons into a shared latent space, enabling unpaired cross-character retargeting~\cite{2020Skeleton-aware}. 
Pose-to-Motion extends this idea to data-scarce scenarios by using static target poses as priors and synthesizes plausible target motions~\cite{zhao2023posetomotion}. 
These studies show that paired motion supervision is not strictly necessary when structural priors and cycle-style objectives are properly designed.

Recent methods in computer graphics also consider geometry, contact and generative modeling more explicitly. 
R2ET uses separate residual modules for skeleton-semantic preservation and geometry-aware correction, reducing artifacts such as interpenetration and missing contacts~\cite{zhang2023}. 
%ReConForM starts from contact semantics and uses transferable mesh key vertices to improve contact accuracy and motion smoothness across diverse character morphologies~\cite{cheynel2025reconformrealtimecontactaware}. 
ReConForM starts from contact semantics and uses a set of transferable key mesh vertices to improve contact accuracy and motion smoothness across diverse character morphologies~\cite{cheynel2025reconformrealtimecontactaware}.
MoReFlow formulates retargeting as unsupervised flow matching between character-specific motion embedding spaces and emphasizes that different application domains prioritize different retargeting objectives, \emph{i.e.} style and visual plausibility for animation, and task-space alignment and executability for robotics~\cite{kim2025moreflowmotionretargetinglearning}.

% Despite these advances, these retargeting methods cannot be directly applied to humanoid robots without additional physics-based modeling. 
% They are commonly evaluated by visual plausibility, contact appearance or geometric consistency, while humanoid robots require executable joint trajectories that satisfy embodiment-specific joint limits, whole-body stability, ground clearance and slip suppression. 
% In addition, they often assume consistent topology, homeomorphic skeletons and transferable mesh correspondences, which does not match the human-to-humanoid setting with different degrees of freedom and mechanical constraints. 
% These differences indicate that previous animation-oriented unpaired retargeting methods must be augmented with robot-specific semantic and physical constraints before applicable to human-to-humanoid motion retargeting.

Despite these advances, these retargeting methods are commonly evaluated by visual plausibility. They cannot be directly applied to humanoid robots where joint trajectories that satisfy embodiment-specific joint limits, whole-body stability, ground clearance and slip suppression are required. 
In addition, they often assume consistent topology, homeomorphic skeletons and transferable mesh correspondences across embodiments, which does not match the human-to-humanoid setting with different degrees of freedom and mechanical constraints. 
Therefore, augmentation of these unpaired retargeting methods regarding robot-specific semantics and physical constraints is necessary for human-to-humanoid motion retargeting.

%% file: tex/3-method.tex
\begin{figure*}[!t] 
    \centering
    \includegraphics[width=1\linewidth]{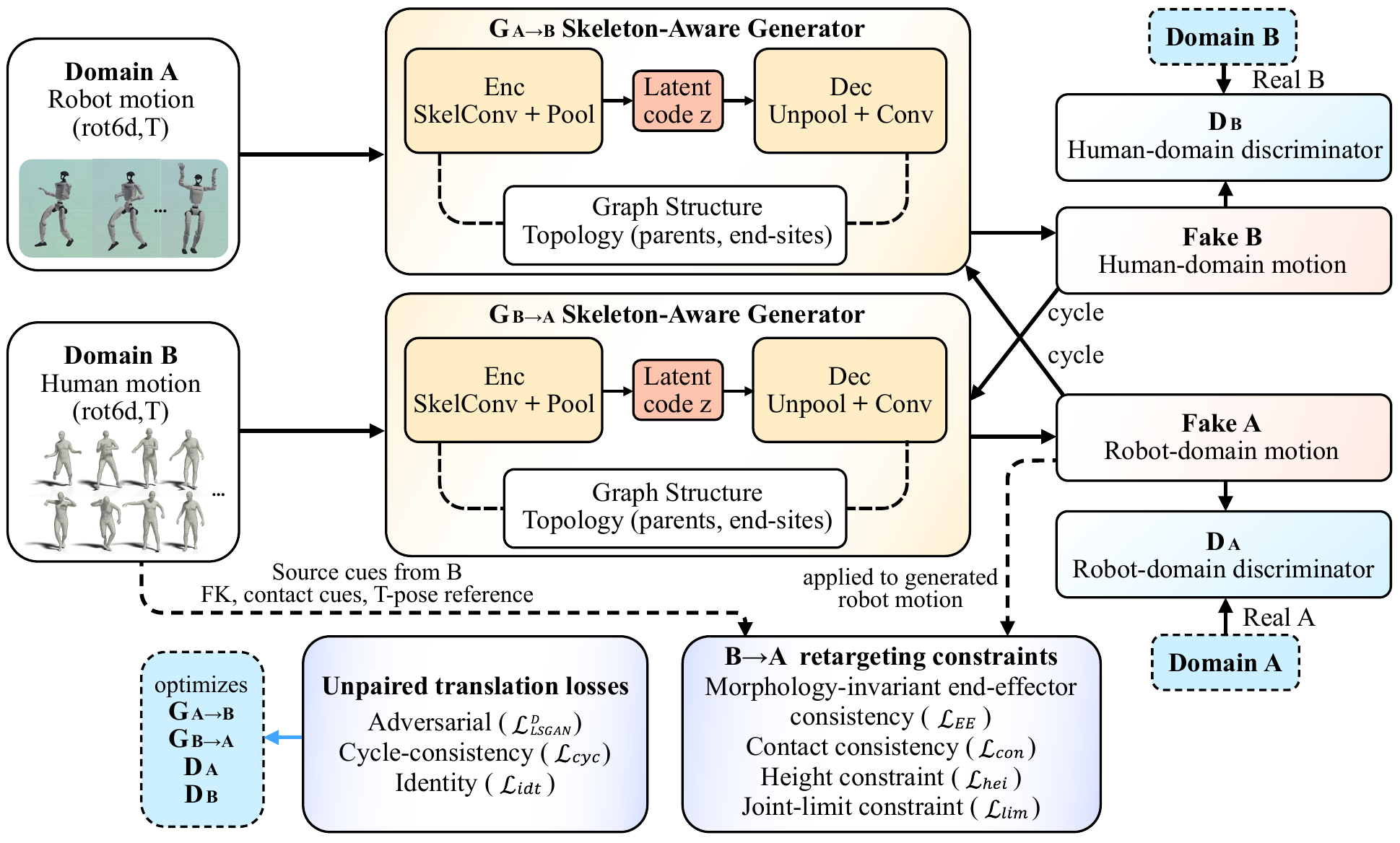}
    % \vspace{-1pt}
    \caption{Overview of the proposed Human2Humanoid framework.}
    \label{fig:overview}
    % \vspace{-15pt}
\end{figure*}

This section presents details of Human2Humanoid. 
The goal is to learn an unpaired mapping $G_{B\to A}$ from the human motion domain $B$ to the humanoid robot motion domain $A$. 
As shown in Fig.~\ref{fig:overview}, the framework comprises two generators $G_{B\to A}$ and $G_{A\to B}$, as well as two discriminators $D_{A}$ and $D_{B}$. 
We assume access to two unpaired motion collections
$\mathcal{Q}_A=\{q_A\}$ and $\mathcal{Q}_B=\{q_B\}$
from domains $A$ and $B$, respectively.
$\mathcal{Q}_A$ and $\mathcal{Q}_B$ have no temporal alignment or frame-wise correspondence.
During training, fixed-length temporal windows are uniformly sampled from each collection
to form mini-batches for adversarial and cycle learning.
To address morphological discrepancies and ensure physical feasibility, we integrate Skeleton-Aware Graph Convolutional Networks (GCNs) and introduce Morphology-Invariant End-Effector Consistency Loss alongside Physics-Aware Feasibility Constraints. 

\subsection{Network Architecture}

\subsubsection{Skeleton-Aware Generator}
To effectively handle skeletal data with hierarchical structures, Skeleton-Aware GCNs~\cite{2020Skeleton-aware} are adopted in our generators, differing from prior methods that treat poses as simple vectors. The generator consists of an encoder, a latent space and a decoder. Instead of flattening joint features, the encoder utilizes \textit{SkeletonConv} layers. This convolution operation explicitly leverages the skeletal topology defined by the adjacency matrix. For a joint $j$, the feature aggregation involves not only itself but also its neighboring joints $\mathcal{N}(j)$, expressed as
\begin{equation}
    f_{\textit{out}}(j) = \sum_{k \in \mathcal{N}(j) \cup \{j\}} W_k \cdot f_{\textit{in}}(k) + b
\end{equation}
where $f_{\textit{in}}(k)$ denotes the input feature of joint $k$, $f_{\textit{out}}(j)$ denotes the output feature of joint $j$ after neighborhood aggregation, 
% $\mathcal{N}(j)$ denotes the set of neighboring joints of joint $j$ in the skeletal graph, 
$W_k$ is a learnable weight for joint $k$, and $b$ is a bias term. The union $\mathcal{N}(j)\cup\{j\}$ indicates that the aggregation includes both the joint and its neighboring joints.
Furthermore, topology-based pooling strategies are employed to compress high-dimensional skeletal features into a low-dimensional latent space, followed by unpooling operations to recover the target topology. This design allows the network to capture local joint correlations and adapt to different kinematic chain structures. The SkeletonConv and topology pooling/unpooling operate on the spatial skeleton graph at each time step.
Temporal coherence is modeled by applying the generator in a sliding window over motion sequences,
which encourages smooth frame-to-frame transitions while preserving local spatial kinematics.

Note that a per-joint correspondence between embodiments is not required. 
The skeleton-aware layers operate on each domain's native kinematic graph, while cross-domain coupling is imposed by the cycle objective and a set of shared semantic end-effectors used in the morphology-invariant constraints.

\subsubsection{Frame-wise Pose Discriminator}
The discriminator operates in a frame-wise manner. %Given a raw input pose sequence represented by per-joint rotation matrices, latent joint features are first extracted using a shared embedding network implemented with $1\times1$ convolutions and LeakyReLU activations. 
Given an input pose sequence represented by per-joint rotation features, latent joint features are first extracted by a joint-shared embedding module implemented with $1\times1$ convolutions and LeakyReLU activations.
Let the resulting pose feature tensor be $F \in \mathbf{R}^{N\times C\times T\times V}$, where $N$ is the mini-batch size, $T$ is the temporal window length, $V$ is the number of joints, and $C=32$ is the latent feature dimension.
To score each frame independently while preserving the joint dimension, $F$ is reshaped to $\mathbf{R}^{(N\cdot T)\times C\times V}$ by folding the temporal dimension into the batch dimension.

Based on this frame-level pose representation, the discriminator contains two complementary branches.
The joint-wise pose branch, denoted as $D_{\textit{each}}$, predicts joint-wise realism scores using joint-specific linear heads, where each head maps the latent feature of one joint from $\mathbf{R}^{C}$ to $\mathbf{R}$.
The full-body pose branch, denoted as $D_{\textit{all}} \in \mathbf{R}^{(N\cdot T)\times 1}$, flattens the full-body latent pose feature and predicts a pose-level coherence score through a Multi-Layer Perceptron.
Before combining the two branches, $D_{\textit{all}}$ is broadcast along the joint dimension, yielding $\widetilde{D}_{\textit{all}} \in \mathbf{R}^{(N\cdot T)\times V}$.
The final joint-wise discriminator output $D \in \mathbf{R}^{(N\cdot T)\times V}$ is obtained by
\begin{equation}
D = w_{\textit{each}} D_{\textit{each}} + \widetilde{D}_{\textit{all}},
\end{equation}
where $w_{\textit{each}}$ controls the contribution of the joint-wise branch.
% , and $\widetilde{D}_{\textit{all}}$ denotes the broadcast version of $D_{\textit{all}}$.

For the adversarial objective, the discriminator output is kept as joint-wise scores instead of being collapsed into a single scalar score. The LSGAN loss is then applied to each joint score independently, and the resulting squared errors are eventually averaged over all joints and frames within the sampled temporal window.

\subsection{Morphology-Invariant End-Effector Consistency Loss}
\label{sec:morphologyLoss}

Humans and humanoid robots often exhibit significant discrepancies in limb lengths and proportions. For instance, human arms are much longer than those of compact robots like Unitree G1. Directly constraining absolute positions in Cartesian space can lead to semantic collapse, \emph{e.g.}, hands failing to reach expected targets. Hence, a Morphology-Invariant End-Effector Consistency Loss is proposed to address this problem.

Instead of aligning absolute coordinates, end-effector trajectories are aligned in a morphology-invariant space defined relative to the rest pose (T-pose).
Specifically, given a source human motion $q_B$ and the generated robot motion $\hat{q}_A = G_{B\to A}(q_B)$, end-effector positions can be computed by forward kinematics and the displacements from the corresponding T-poses can be compared.
Let $\mathcal{E}$ denote the set of shared semantic end-effectors
(left/right hands and feet) defined in both domains, and let
$\operatorname{FK}_k(q,t)$ return the position of end-effector $k$
at frame $t$.
The corresponding displacements are normalized by the embodiment-specific constant body scale $S$ computed from the rest pose to avoid variation across sequences within the same embodiment:

% {\small
% \begin{equation}
% \begin{aligned}
% \mathcal{L}_{EE}
% &=
% \mathbf{E}\!\Bigg[
% \frac{1}{T|\mathcal{E}|}
% \sum_{t=1}^{T}\sum_{k\in\mathcal{E}}
% \Bigg\|
% \frac{\operatorname{FK}_k(q_B,t)-FK_k(q_B^{T})}{S_B}
% \\[-1mm]
% &\hspace{30mm}
% -
% \frac{FK_k(\hat{q}_A,t)-FK_k(q_A^{T})}{S_A}
% \Bigg\|_2^2
% \Bigg]
% \end{aligned}
% \end{equation}
% }

\begin{equation}
\begin{aligned}
\mathcal{L}_{\textit{EE}}
&=
\mathbf{E}\!\Bigg[
\frac{1}{T|\mathcal{E}|}
\sum_{t=1}^{T}\sum_{k\in\mathcal{E}}
\Bigg\|
\frac{\operatorname{FK}_k(q_B,t)-\operatorname{FK}_k(q_B^{\mathit{Tpose}})}{S_B}
\\
&\hspace{20mm}
-
\frac{\operatorname{FK}_k(\hat{q}_A,t)-\operatorname{FK}_k(q_A^{\mathit{Tpose}})}{S_A}
\Bigg\|_2^2
\Bigg],
\end{aligned}
\end{equation}

% \noindent where $\mathbf{E}[\cdot]$ denotes expectation over sampled motion windows, and $q_B^{T}$ and $q_A^{T}$ denote the T-pose configurations in the human and robot domains, respectively.
\noindent where $\mathbf{E}[\cdot]$ denotes expectation over the sampled temporal window, and
$q_B^{\mathit{Tpose}}$ and $q_A^{\mathit{Tpose}}$ denote the T-pose configurations
in the human and robot domains, respectively.
% , and $S_B$ and $S_A$ are the corresponding embodiment-specific body scales computed from the rest pose.

\subsection{Physics-Aware Feasibility Constraints}
\label{sec:physicsLoss}

%Unsupervised retargeting often suffers from physically infeasible artifacts such as foot skating, base floating or violation of mechanical limits. To enhance deployability on real robots, explicit contact constraints and joint limit penalties are introduced during training. Unless otherwise specified, the following constraints are applied to the generation of robot motion $\hat{q}_A$ through $G_{B\to A}$.
Unsupervised retargeting often suffers from physically infeasible artifacts such as foot skating, base floating and violation of mechanical limits. To enhance deployability on real robots, explicit foot contact, foot height and joint limit constraints are introduced during training. Unless otherwise specified, the following constraints are applied to the generation of robot motion $\hat{q}_A$ through $G_{B\to A}$.

\subsubsection{Foot Contact Constraint}
A binary contact indicator $c_{B}^{(m)}$ is inferred from the source human motion (domain $B$) using the foot speed.
Let $m \in \mathcal{F}$ index the feet and let $t$ denote time, thus
\begin{equation}
c_{B}^{(m)}(t) = \mathbf{1}\!\left(\left\| v_{B}^{(m)}(t) \right\|_2 < \tau \right),
\end{equation}
where $v_{B}^{(m)}(t)$ is the source foot velocity and $\tau$ is a small threshold.
Using this contact gate, foot skating is suppressed in the generated robot motion in domain $A$ by penalizing the target foot velocity as

% {\small
% \begin{equation}
%     \mathcal{L}_{\textit{con}}
%     =
%     \frac{1}{\sum_{t}\sum_{m\in\mathcal{F}} c_{B}^{(m)}(t) + \epsilon}
%     \sum_{t}\sum_{m\in\mathcal{F}}
%     c_{B}^{(m)}(t)\, \left\| \frac{v_{A}^{(m)}(t)}{S_A} \right\|_2,
% \end{equation}
% }

\begin{equation}
    \mathcal{L}_{\textit{con}}
    =
    \frac{\sum_{t}\sum_{m\in\mathcal{F}}
    c_{B}^{(m)}(t)\, \left\| \frac{v_{A}^{(m)}(t)}{S_A} \right\|_2}{\sum_{t}\sum_{m\in\mathcal{F}} c_{B}^{(m)}(t) + \epsilon},
\end{equation}

\noindent where $\epsilon$ is a small constant for numerical stability, preventing division by zero when no source foot is detected in contact.

\subsubsection{Foot Height Constraint}

To further improve physical plausibility during stance, foot hovering is also penalized in the generated robot motion.
Let $h_{A}^{(m)}(t)$ and $h_{B}^{(m)}(t)$ denote the vertical heights of robot and human foot $m$ at time $t$, respectively.
For each domain, a nominal foot contact height is pre-computed from the T-pose, denoted as $h_{A,m}^{\textit{ref}}$ and $h_{B,m}^{ref}$.
A stance-weight mask is then computed using the source domain data as
\begin{equation}
    w^{(m)}(t) \;=\; c_{B}^{(m)}(t)\,\cdot\, \mathbf{1}\!\left( h_{B}^{(m)}(t) < h_{B,m}^{ref} \right).
\end{equation}
This additional height-based filter removes spurious low-velocity detections. 
% by requiring the source foot to be no higher than its nominal grounded height in the T-pose.
Consequently, an anti-floating hinge penalty is applied on the target foot height as
% {\small
% \begin{equation}
% \begin{aligned}
% \mathcal{L}_{\textit{hei}}
% &=
% \frac{1}{\sum_{t}\sum_{m\in\mathcal{F}} w^{(m)}(t) + \epsilon}
% \sum_{t}\sum_{m\in\mathcal{F}}
% w^{(m)}(t)
% \\[-1mm]
% &\hspace{10mm}
% \left[
% \frac{\mathrm{ReLU}\!\big(h_{A}^{(m)}(t) - h_{A,m}^{ref}\big)}{S_A}
% \right]^2 .
% \end{aligned}
% \end{equation}
% }
\begin{equation}
\mathcal{L}_{\textit{hei}}
=
\frac{\sum_{t}\sum_{m\in\mathcal{F}}
w^{(m)}(t) \left[
\frac{\mathrm{ReLU}\!\big(h_{A}^{(m)}(t) - h_{A,m}^{\textit{ref}}\big)}{S_A}
\right]^2}{\sum_{t}\sum_{m\in\mathcal{F}} w^{(m)}(t) + \epsilon},
\end{equation}
where $h_{A,m}^{\textit{ref}}$ encourages the target foot to remain close to its nominal grounded height during source-inferred contact.

\subsubsection{Joint Limit Constraints}
Since humanoid robots have strict mechanical ranges of motion, generated motions that exceed these limits may damage hardware or trigger emergency stops.
Therefore, a joint limit loss is introduced to penalize the predicted robot joint angles $\hat{q}^{A}$ falling outside the range $[q_{min}, q_{max}]$ as
\begin{equation}
\begin{aligned}
\mathcal{L}_{\textit{lim}}
&=
\sum_{t}\sum_{j}
\Big(
\big\|\mathrm{ReLU}(\hat{q}_{A,t,j}-q_{\max,j})\big\|^{2}
\\
&\hspace{10mm}
+
\big\|\mathrm{ReLU}(q_{\min,j}-\hat{q}_{A,t,j})\big\|^{2}
\Big),
\end{aligned}
\end{equation}
where $j$ is the joint index. 
% This ensures the generated motions remain within the feasible configuration space of the robot.

% \subsection{Full Objective Function}
% The training of Human2Humanoid is driven by a composite objective. In addition to the retargeting-specific losses, we also employ adversarial losses to match data distributions, cycle-consistency losses to ensure motion reversibility, and identity losses for regularization. 
% To stabilize training and improve generation quality, the \textbf{Least Squares GAN (LSGAN)} objective~\cite{mao2017} is adopted.
% Let $q_A \sim P_A$ and $q_B \sim P_B$ denote unpaired motion samples from the robot and human domains, respectively.
\subsection{Full Objective Function}
The training of Human2Humanoid is driven by a composite objective. In addition
to the retargeting-specific losses introduced in \ref{sec:morphologyLoss} and \ref{sec:physicsLoss}, we also employ adversarial losses to match data
distributions, cycle-consistency losses to ensure motion reversibility, and
identity losses for regularization. 

To stabilize training and improve generation
quality, the Least Squares GAN (LSGAN) objective~\cite{mao2017} is
adopted. Let $P_A$ and $P_B$ denote the empirical motion distributions induced
by the unpaired collections $\mathcal{Q}_A$ and $\mathcal{Q}_B$, respectively.
Thus, samples $q_A \sim P_A$ and $q_B \sim P_B$ are drawn during training.
For each sampled window, the discriminators output scores for each joint at each frame. We therefore implement the LSGAN objective element-wise and average the squared errors over all joints and frames. Let $D_{A,t,j}(q)$ denote the score of discriminator $D_A$ for joint $j$ at frame $t$, and analogously for $D_B$, thus

% {\small
% \begin{align}
% \mathcal{L}^{D}_{\textit{LSGAN}}
% &=
% \mathbf{E}_{q_A \sim P_A}\!\left[
% \frac{1}{TV}\sum_{t=1}^{T}\sum_{j=1}^{V}\big(D_{A,t,j}(q_A)-1\big)^2
% \right]
% \nonumber\\
% &\quad+
% \mathbf{E}_{q_B \sim P_B}\!\left[
% \frac{1}{TV}\sum_{t=1}^{T}\sum_{j=1}^{V}\big(D_{A,t,j}(G_{B\to A}(q_B))\big)^2
% \right]
% \nonumber\\
% &\quad+
% \mathbf{E}_{q_B \sim P_B}\!\left[
% \frac{1}{TV}\sum_{t=1}^{T}\sum_{j=1}^{V}\big(D_{B,t,j}(q_B)-1\big)^2
% \right]
% \nonumber\\
% &\quad+
% \mathbf{E}_{q_A \sim P_A}\!\left[
% \frac{1}{TV}\sum_{t=1}^{T}\sum_{j=1}^{V}\big(D_{B,t,j}(G_{A\to B}(q_A))\big)^2
% \right].
% \end{align}
% }
\begin{equation}
\begin{aligned}
\mathcal{L}^{D}_{\textit{LSGAN}}
&=
\mathbf{E}_{q_A \sim P_A}\!\left[
\frac{1}{TV}\sum_{t=1}^{T}\sum_{j=1}^{V}\big(D_{A,t,j}(q_A)-1\big)^2
\right]
\\
&\hspace{-5mm}+
\mathbf{E}_{q_B \sim P_B}\!\left[
\frac{1}{TV}\sum_{t=1}^{T}\sum_{j=1}^{V}\big(D_{A,t,j}(G_{B\to A}(q_B))\big)^2
\right]
\\
&\hspace{-5mm}+
\mathbf{E}_{q_B \sim P_B}\!\left[
\frac{1}{TV}\sum_{t=1}^{T}\sum_{j=1}^{V}\big(D_{B,t,j}(q_B)-1\big)^2
\right]
\\
&\hspace{-5mm}+
\mathbf{E}_{q_A \sim P_A}\!\left[
\frac{1}{TV}\sum_{t=1}^{T}\sum_{j=1}^{V}\big(D_{B,t,j}(G_{A\to B}(q_A))\big)^2
\right].
\end{aligned}
\end{equation}
\noindent On the other hand, the generators are trained to fool the discriminators by regressing generated samples to $1$ as

{\small
\begin{align}
\mathcal{L}^{G}_{\textit{LSGAN}}
&=
\mathbf{E}_{q_B \sim P_B}\!\left[
\frac{1}{TV}\sum_{t=1}^{T}\sum_{j=1}^{V}\big(D_{A,t,j}(G_{B\to A}(q_B))-1\big)^2
\right]
\nonumber\\
&\hspace{-5mm}+
\mathbf{E}_{q_A \sim P_A}\!\left[
\frac{1}{TV}\sum_{t=1}^{T}\sum_{j=1}^{V}\big(D_{B,t,j}(G_{A\to B}(q_A))-1\big)^2
\right].
\end{align}
}

\noindent The cycle-consistency and identity losses are defined using the $L_1$ norm as
\begin{align}
\mathcal{L}_{\textit{cyc}} &=
\mathbf{E}_{q_B}\!\left[\left\|G_{A\to B}(G_{B\to A}(q_B)) - q_B\right\|_1\right] \nonumber\\
&\quad+
\mathbf{E}_{q_A}\!\left[\left\|G_{B\to A}(G_{A\to B}(q_A)) - q_A\right\|_1\right], \\
\mathcal{L}_{\textit{idt}} &=
\mathbf{E}_{q_A}\!\left[\left\|G_{B\to A}(q_A) - q_A\right\|_1\right] \nonumber\\
&\quad+
\mathbf{E}_{q_B}\!\left[\left\|G_{A\to B}(q_B) - q_B\right\|_1\right].
\end{align}
Therefore, the full generator objective can be written as
\begin{equation}
\begin{aligned}
\mathcal{L}_{G}
&=
\lambda_{\textit{GAN}}\mathcal{L}^{G}_{\textit{LSGAN}}
+\lambda_{\textit{cyc}}\mathcal{L}_{\textit{cyc}}
+\lambda_{\textit{idt}}\mathcal{L}_{\textit{idt}}
\\
&\quad
+\lambda_{\textit{EE}}\mathcal{L}_{\textit{EE}}
+\lambda_{\textit{con}}\mathcal{L}_{\textit{con}}
+\lambda_{\textit{hei}}\mathcal{L}_{\textit{hei}}
% \\
% &\quad
+\lambda_{\textit{lim}}\mathcal{L}_{\textit{lim}} ,
\end{aligned}
\end{equation}
where the $\lambda$ terms are scalar weights balancing the corresponding objectives.
Eventually, the generators are optimized by minimizing $\mathcal{L}_{G}$, while the discriminators are optimized by minimizing $\mathcal{L}^{D}_{\textit{LSGAN}}$.

%% file: tex/4-exp.tex
\label{sec:experiments}
\subsection{Experimental Setup}
\label{sec:exp_setup}

To evaluate Human2Humanoid under substantial differences in skeletal topology, scale and degrees of freedom, we train and evaluate the model on two unpaired motion domains. The human domain is constructed from Motion-X~\cite{Motion-X}, which provides large-scale 3D full-body motion annotations in the SMPL-X format and covers a broad range of motion semantics and scenarios. The robot domain is constructed from the Unitree G1 subset of PHUMA (Physically-Grounded Humanoid Locomotion Dataset)~\cite{PHUMA}.
% To evaluate the ability of Human2Humanoid to perform unpaired motion retargeting across heterogeneous embodiments with substantial differences in skeletal topology, scale and degrees of freedom, 
% the source human motion uses Motion-X~\cite{Motion-X} to provide large-scale 3D full-body motion annotations in the SMPL-X format and cover a broad range of motion semantics and scenarios, and the target robot motion uses the Unitree G1 subset inside PHUMA (Physically-Grounded Humanoid Locomotion Dataset)~\cite{PHUMA}. 
% the proposed model is trained and tested on two large motion domains: a human motion domain $\mathcal{B}$ and a robot motion domain $\mathcal{A}$.
% For the source domain, \textbf{Motion-X}~\cite{Motion-X} is used to provide large-scale 3D full-body motion annotations in the SMPL-X format and cover a broad range of motion semantics and scenarios.
% For the target domain, the Unitree G1 subset inside \textbf{PHUMA (Physically-Grounded Humanoid Locomotion Dataset)}~\cite{PHUMA} is selected as the robot data.
PHUMA applies physics-based filtering and optimization to reduce common artifacts such as foot skating and ground penetration, providing a target distribution that better reflects executable robot motions.

\subsubsection{Preprocessing}
Since Motion-X and PHUMA differ in original frame rates and motion representations, a unified preprocessing is applied.
%Sequences in both domains are resampled to $30$ Hz and each long sequence is segmented into fixed-length temporal clips using a sliding window of length $T=64$, with stride $s=1$.
Sequences in both domains are resampled to $30$ Hz. Each long sequence is segmented into fixed-length temporal clips of $64$ frames using a sliding window with a one-frame stride, so adjacent clips overlap by $63$ frames.
Then, pose parameters in Motion-X are converted to joint rotation matrices, and G1 joint data in PHUMA are also converted into rotation-matrix form, augmented with root-related features.
For the root, the absolute world translations are not used. Instead, the frame-wise differences are used to compute the root linear velocity as input, yielding translation invariance to absolute position.
Meanwhile, the root global rotation matrix is kept as input and no yaw-normalization (de-heading) is applied during training.
Train/test splits are performed independently within each domain. No temporal synchronization or semantic correspondence is used during training, strictly adhering to the unpaired learning assumption.

\subsubsection{Baselines}
Comparison against representative robot retargeting baselines for Unitree G1 is conducted, including two optimization-based pipelines and one industrial-quality reference.
(i) PHC retargeting~\cite{phc} is an optimization-based SMPL-to-humanoid retargeting pipeline commonly used to produce executable reference motions for humanoid robots.
(ii) GMR~\cite{gmr} is an optimization-based retargeting method with a two-stage constrained solver and feasibility-oriented constraints.
(iii) Unitree Retarget (closed-source reference) is a set of reference motions for G1 humanoid robot produced by an industrial pipeline. It is used as a high-quality reference rather than a learnable baseline.
% in real-robot deployment, rather than as a learnable model.

\subsection{Evaluation Metrics}
\label{sec:metrics}

% The retargeted motions are evaluated from two aspects: (i) \textbf{downstream controllability} under a fixed tracking policy in simulation, and (ii) \textbf{real-robot deployability} on Unitree G1.
% In addition, offline physical feasibility diagnostics (FS/GP) are reported to interpret common failure modes.
The retargeted motions are evaluated in simulation on Unitree G1 under a fixed tracking policy from two aspects: (i) downstream controllability, and (ii) physical feasibility.

\subsubsection{Downstream Controllability}
\label{sec:sim_tracking}

% A generic humanoid motion tracking policy is used to evaluate whether retargeted motions are easy to track.
The pretrained Unitree G1 tracking policy from the open-source humanoid-general-motion-tracking project~\cite{chen2025gmt} is used to evaluate whether retargeted motions are easy to track. The tracking policy is kept fixed to help evaluation focus on the controllability of reference motion and enables a fair comparison across retargeting methods.
The resulting reference trajectory from each retargeted motion clip is fed into the project's simulation evaluation pipeline and the policy is rolled out under the same configuration.
The corresponding evaluation metrics include
(i) Success Rate (SR), indicating the fraction of clips whose rollouts complete without falling/termination under the project's default termination conditions, and
(ii) Tracking Error (TE), representing the average tracking error computed using the project's default definition and logged quantities.
% Using a fixed pretrained policy helps focus on the effect of reference motion quality on controllability and enables a fair comparison across retargeting methods.

\subsubsection{Physical Feasibility}
\label{sec:feasibility}

Unless otherwise specified, physical feasibility metrics are computed on the test split using the same forward-kinematics routines and coordinate conventions as in training. The following two evaluation metrics are used:

% \paragraph{Foot Skating (FS)}
% FS measures the fraction of foot-frame pairs in which the robot foot is close to ground while exhibiting excessive horizontal slip.
% Following the simulator-based diagnostic pipeline, foot contact is inferred from the minimum height of foot geometry:
\paragraph{Foot Skating (FS)}
FS measures the fraction of frames where the robot exhibits excessive horizontal slip during ground contact.
Ground contact is inferred from the minimum height of the foot geometry as
\begin{equation}
c_{A}^{(m)}(t)=
\mathbf{1}\!\left(h_{A}^{(m)}(t)<h_{\mathit{contact}}\right),
\end{equation}
where $m\in\mathcal{F}$ indexes the feet, $\mathcal{F}$ denotes the foot set, and $h_{\mathit{contact}}$ is a contact height threshold.
A skating event is incurred if the foot's horizontal velocity exceeds a tolerance threshold $\tau_{\mathrm{tol}}$ during contact.
Therefore, FS is computed as
\begin{equation}
\mathrm{FS}
=
\frac{1}{T|\mathcal{F}|}
\sum_{t=1}^{T}\sum_{m\in\mathcal{F}}
c_{A}^{(m)}(t)\cdot
\mathbf{1}\!\left(\left\|v_{A,xy}^{(m)}(t)\right\|_2>\tau_{\mathrm{tol}}\right),
\end{equation}
with $h_{\mathrm{contact}}=0.0025\,\mathrm{m}$ and $\tau_{\mathrm{tol}}=0.30\,\mathrm{m/s}$.

% \paragraph{Foot Skating Ratio (FS)}
% FS measures the fraction of frames where the foot in source motion is inferred to be in ground contact while the robot foot exhibits excessive horizontal slip.
% A binary contact indicator is first inferred from the source (human) motion using a velocity threshold $\tau$:
% \begin{equation}
% c_{B}^{(m)}(t)=\mathbf{1}\!\left(\left\|v_{B}^{(m)}(t)\right\|_2<\tau\right),
% \end{equation}
% where $m\in\mathcal{F}$ indexes foot end-effectors and $\mathcal{F}$ denotes the foot set.
% A skating event is counted if $c_{B}^{(m)}(t)=1$ and the target horizontal speed $\|v_{A,xy}^{(m)}(t)\|_2$ of robot foot exceeds a tolerance threshold $\tau_{\mathrm{tol}}$.
% The FS is then
% \begin{equation}
% \mathrm{FS}
% =
% \frac{1}{T|\mathcal{F}|}
% \sum_{t=1}^{T}\sum_{m\in\mathcal{F}}
% c_{B}^{(m)}(t)\cdot
% \mathbf{1}\!\left(\left\|v_{A,xy}^{(m)}(t)\right\|_2>\tau_{\mathrm{tol}}\right).
% \end{equation}
% Unless otherwise specified, we set the contact threshold to $\tau=0.003$ and the skating tolerance threshold to $\tau_{\mathrm{tol}}=0.02\,\mathrm{m/s}$, with units consistent with preprocessing.
% %We use the same threshold configuration as in the data loader ($\tau=0.003$ and $\tau_{\mathrm{tol}}=0.02\,\mathrm{m/s}$, with units consistent with preprocessing).

\paragraph{Ground Penetration (GP)}
GP measures the average ground penetration depth of the robot feet and is defined as
% below the ground plane ($h<0$).
% Let $h_{A}^{(m)}(t)$ be the vertical coordinate of robot foot end-effector $m$ at frame $t$, thus
\begin{equation}
\mathrm{GP}
=
\frac{1}{T|\mathcal{F}|}
\sum_{t=1}^{T}\sum_{m\in\mathcal{F}}
\max\!\left(0,\,-h_{A}^{(m)}(t)\right).
\end{equation}

% \paragraph{Joint Limit Violation (JLV).}
% JLV is the fraction of frames in which any joint exceeds the mechanical limits specified by the robot URDF (or soft limits).
% A frame $t$ is counted as a violation if there exists a joint $j$ such that $\hat{q}^{A}_{t,j}\notin[q_{\min,j},q_{\max,j}]$:
% \begin{equation}
% \mathrm{JLV}
% =
% \frac{1}{T}
% \sum_{t=1}^{T}
% \mathbf{1}\!\left(
% \exists j,\ 
% \hat{q}^{A}_{t,j}<q_{\min,j}\ \text{or}\ \hat{q}^{A}_{t,j}>q_{\max,j}
% \right).
% \end{equation}
% Unless otherwise specified, JLV is computed on the raw generator outputs before the final clipping-to-limits step during motion sequence export.

\subsection{Evaluation Results}
\label{sec:quant}

\begin{table*}[t]
\centering
\small
\caption{Per-motion quantitative comparison of retargeting performance from human to Unitree G1. G, P, U and O denote GMR, PHC, Unitree Retarget and Ours (Human2Humanoid), respectively. The best and second-best results are highlighted in bold and with underscores, respectively.}
\label{tab:quantitative}
\resizebox{\textwidth}{!}{%
\renewcommand{\arraystretch}{1.1}

\begin{tabular}{lcccccccccccccccc}
\toprule
\multirow{2}{*}{\textbf{Motion}}
& \multicolumn{4}{c}{\textbf{SR (\%) $\uparrow$}}
& \multicolumn{4}{c}{\textbf{TE $\downarrow$}}
& \multicolumn{4}{c}{\textbf{FS (\%) $\downarrow$}}
& \multicolumn{4}{c}{\textbf{GP (cm) $\downarrow$}} \\
\cmidrule(lr){2-5} \cmidrule(lr){6-9} \cmidrule(lr){10-13} \cmidrule(lr){14-17}
& G & P & U & \textbf{O}
& G & P & U & \textbf{O}
& G & P & U & \textbf{O}
& G & P & U & \textbf{O} \\
\midrule
Walk1 & \textbf{95.0} & 45.0 & \textbf{95.0} & \underline{90.0} & \underline{0.12} & 0.16 & 0.16 & \textbf{0.11} & 23.7 & \textbf{3.5} & 37.6 & \underline{17.7} & 0.55 & \underline{0.24} & 1.80 & \textbf{0.03} \\
Walk2 & \textbf{90.0} & 25.0 & \textbf{90.0} & \underline{85.0} & \underline{0.13} & 0.17 & 0.16 & \textbf{0.11} & 19.3 & \textbf{4.0} & 30.3 & \underline{14.1} & 0.32 & \underline{0.30} & 0.97 & \textbf{0.02} \\
Walk3 & 85.0 & 35.0 & \underline{90.0} & \textbf{100.0} & \underline{0.15} & 0.17 & 0.17 & \textbf{0.10} & 13.9 & \textbf{1.9} & \underline{9.5} & 11.5 & 0.19 & 0.13 & \underline{0.10} & \textbf{0.00} \\
Hop1 & 70.0 & 10.0 & \textbf{90.0} & \underline{85.0} & \underline{0.16} & \underline{0.16} & 0.17 & \textbf{0.11} & \textbf{0.0} & \textbf{0.0} & 3.5 & \underline{0.2} & \textbf{0.00} & \textbf{0.00} & \textbf{0.00} & \textbf{0.00} \\
Jump1 & \textbf{90.0} & 5.0 & \textbf{90.0} & \underline{85.0} & \underline{0.14} & 0.18 & 0.18 & \textbf{0.12} & 21.8 & \textbf{8.4} & 33.0 & \underline{16.1} & \underline{0.56} & 0.64 & 1.69 & \textbf{0.03} \\
Box1 & \textbf{95.0} & \underline{0.0} & \underline{0.0} & \textbf{95.0} & \underline{0.14} & 0.18 & 0.21 & \textbf{0.10} & 0.2 & \textbf{0.0} & \textbf{0.0} & \underline{0.1} & \textbf{0.00} & \textbf{0.00} & \textbf{0.00} & \underline{0.04} \\
Box2 & \textbf{80.0}& 0.0 & 0.0 & \underline{70.0} & \textbf{0.13} & \underline{0.17} & 0.19 & \textbf{0.13} & \underline{0.3} & \textbf{0.0} & 2.5 & 0.7 & \textbf{0.00} & \textbf{0.00} & \textbf{0.00} & \underline{0.55} \\
Crouch1 & \textbf{50.0} & \underline{0.0} & \underline{0.0} & \textbf{50.0} & \underline{0.29} & 0.92 & 0.56 & \textbf{0.21} & \textbf{0.0} & \underline{0.3} & \textbf{0.0} & \textbf{0.0} & \textbf{0.00} & \textbf{0.00} & \textbf{0.00} & \textbf{0.00} \\
Backward1 & 90.0& \underline{95.0} & \textbf{100.0} & \textbf{100.0}& \textbf{0.10} & 0.14 & 0.12 & \underline{0.11} & 2.2 & \textbf{0.0} & 9.8 & \underline{1.3} & \textbf{0.00} & \textbf{0.00} & \textbf{0.00} & \textbf{0.00} \\
Backward2 & 85.0& 35.0 & \underline{90.0} & \textbf{95.0}& \underline{0.12} & 0.16 & 0.13 & \textbf{0.11} & \underline{0.4} & \textbf{0.0} & 3.6 & \textbf{0.0} & \textbf{0.00} & \textbf{0.00} & \textbf{0.00} & \textbf{0.00} \\
Backward3 & \textbf{100.0} & \underline{75.0} & \textbf{100.0} & \textbf{100.0} & \underline{0.09} & 0.14 & 0.11 & \textbf{0.08} & \underline{0.8} & \textbf{0.0} & 8.0 & \textbf{0.0} & \textbf{0.00} & \textbf{0.00} & \textbf{0.00} & \textbf{0.00} \\
Turn1 & \textbf{100.0} & 35.0 & 90.0 & \underline{95.0} & \underline{0.13} & 0.15 & 0.17 & \textbf{0.09} & \underline{0.2} & \textbf{0.0} & \underline{0.2} & \textbf{0.0} & \textbf{0.00} & \textbf{0.00} & \textbf{0.00} & \textbf{0.00} \\
Stand1 & \textbf{100.0} & 65.0 & \underline{90.0} & \textbf{100.0}& \textbf{0.12} & \underline{0.15} & 0.17 & \textbf{0.12} & 5.5 & \underline{0.3} & 6.5 & \textbf{0.0} & \textbf{0.00} & \underline{0.16} & \textbf{0.00} & \textbf{0.00} \\
\midrule
Avg.
& \underline{86.9} & 32.7 & 71.2 & \textbf{88.5}
& \underline{0.14} & 0.22 & 0.19 & \textbf{0.12}
& 6.8 & \textbf{1.4} & 11.1 & \underline{4.7}
& 0.12 & \underline{0.11} & 0.35 & \textbf{0.05} \\
\hline
\end{tabular}%
}
\end{table*}

% The proposed Human2Humanoid is compared with PHC retargeting, GMR, and Unitree Retarget in terms of downstream controllability (SR and TE) and physical feasibility (FS and GP).
The quantitative results are summarized in Table~\ref{tab:quantitative}.
%Note that the results from the optimization-based baselines and the closed-source reference are reported as values without variances.
% All quantitative results are reported as values without variances under the same fixed evaluation protocol.

\subsubsection{General Discussion}

Under the same tracking policy, Human2Humanoid achieves the best TE among the compared methods, indicating that its generated reference motions are generally easier to track. GMR obtains competitive SR, but its average TE, FS and GP remain higher than those of Human2Humanoid, suggesting that stronger SR does not fully eliminate contact-related artifacts. PHC achieves the lowest average FS, but exhibits substantially lower SR on several challenging motions, indicating that contact feasibility alone does not necessarily lead to reliably executable reference motions. The Unitree Retarget provides an industrial-quality reference, yet Human2Humanoid achieves higher average SR and lower TE under the same evaluation protocol. Besides, Human2Humanoid achieves the lowest average GP among all compared methods, indicating better ground penetration suppression. Its average FS is also lower than those of GMR and Unitree Retarget. Overall, these results show that Human2Humanoid provides a more trackable reference motion distribution than the compared baselines.

\subsubsection{Failure cases of the optimization-based baselines}
\label{sec:failure_cases}
\begin{figure}[t] 
    \centering
    \subfloat[]{
    \includegraphics[width=0.47\columnwidth]{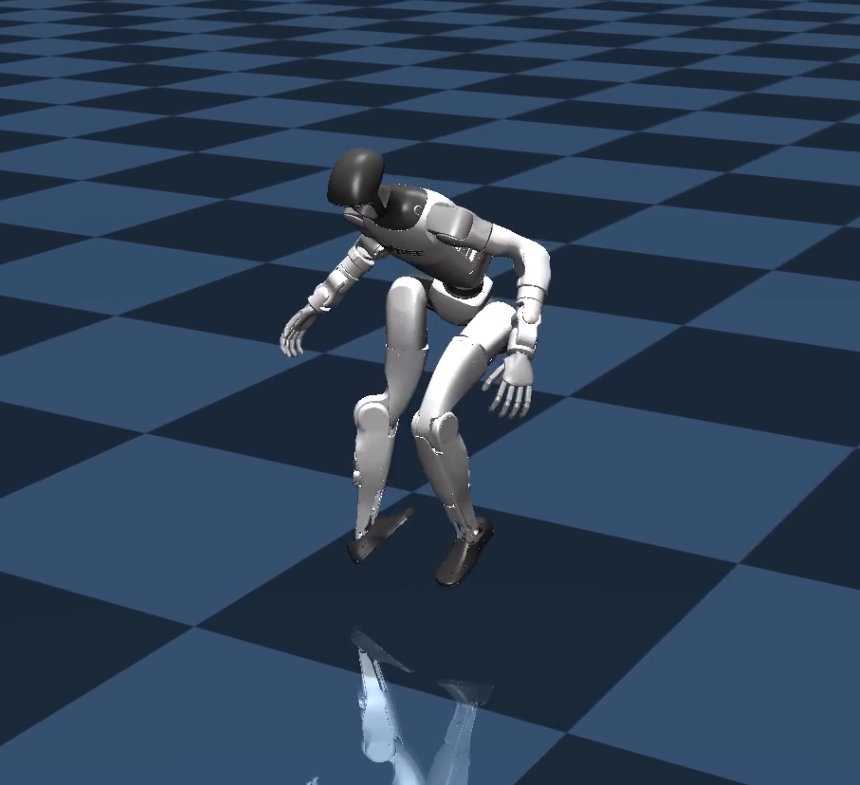}
    }
    \subfloat[]{
    \includegraphics[width=0.47\columnwidth]{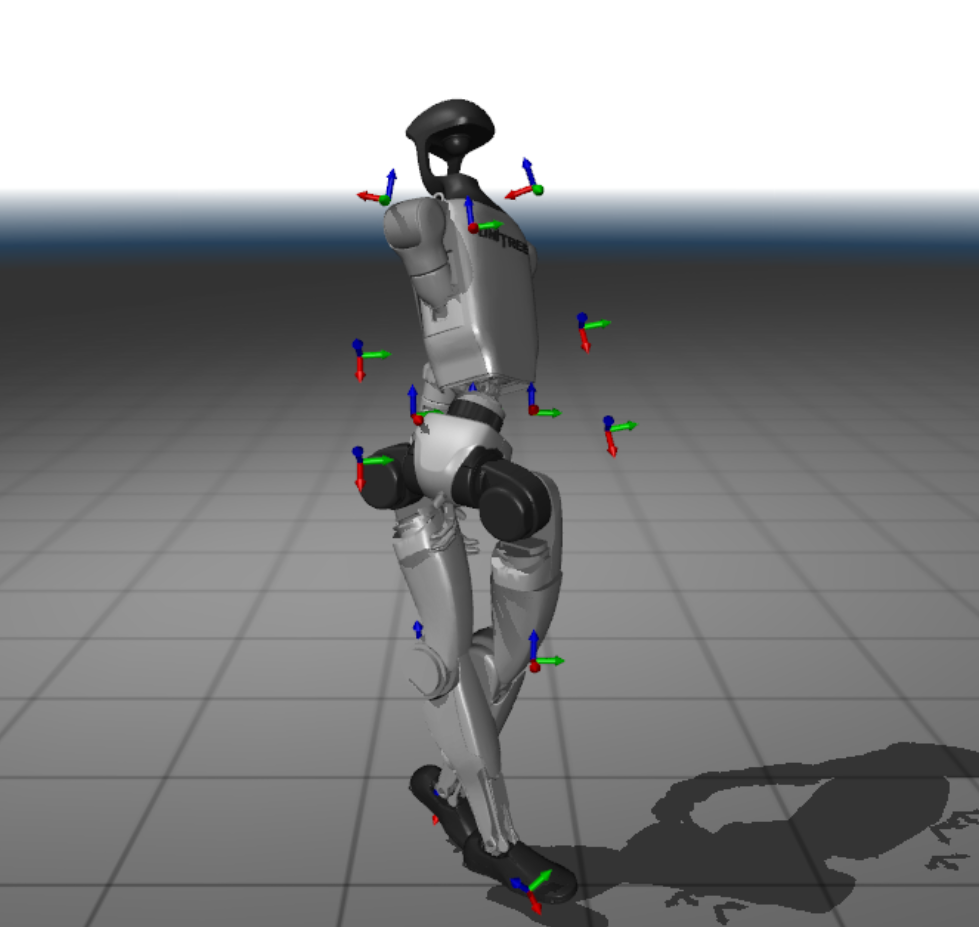}
    }
    \caption{Representative failure cases of the optimization-based baselines on challenging motions.
    (a) PHC shows joint-limit induced discontinuity.
    (b) GMR shows jitter without per-motion tuning.}
    \label{fig:failure_cases}    
\end{figure}

Although optimization-based pipelines can enforce explicit constraints, it is observed that they may still produce catastrophic artifacts on hard motions, especially near joint limits or under large morphology mismatches.
Fig.~\ref{fig:failure_cases} shows two representative examples.
In PHC, when the motion approaches the robot's joint limits, the solver may switch to a different local solution branch, causing untrackable joint “snapping”.
In GMR, some sequences require non-trivial per-clip parameter tuning to avoid noticeable jitter.
Such failures are difficult to fully filter out without manual inspection, while the proposed method can reduce these catastrophic cases.

% \begin{figure}[t]
%   \centering
%   \begin{minipage}[t]{0.48\linewidth}
%     \centering
%     \includegraphics[width=\linewidth]{imgs/fail_phc.png}
%     \vspace{-5mm}
%     \caption*{(a)}
%     % \caption*{(a) PHC: joint-limit induced discontinuity.}
%   \end{minipage}\hfill
%   \begin{minipage}[t]{0.48\linewidth}
%     \centering
%     \includegraphics[width=\linewidth]{imgs/fail_gmr.png}
%     \vspace{-5mm}
%     \caption*{(b)}
%     % \caption*{(b) GMR: jitter without per-motion tuning.}
%   \end{minipage}
%   \vspace{-1mm}
%   \caption{Representative failure cases of the optimization-based baselines on challenging motions. These artifacts can yield untrackable reference trajectories and are hard to fully eliminate via manual filtering/tuning at dataset scale. (a) PHC: joint-limit induced discontinuity. (b) GMR: jitter without per-motion tuning.}
%   \label{fig:failure_cases}
%   \vspace{-2mm}
% \end{figure}

\subsubsection{Qualitative comparison}
\begin{figure}[t] 
    \centering
    \includegraphics[width=1\linewidth]{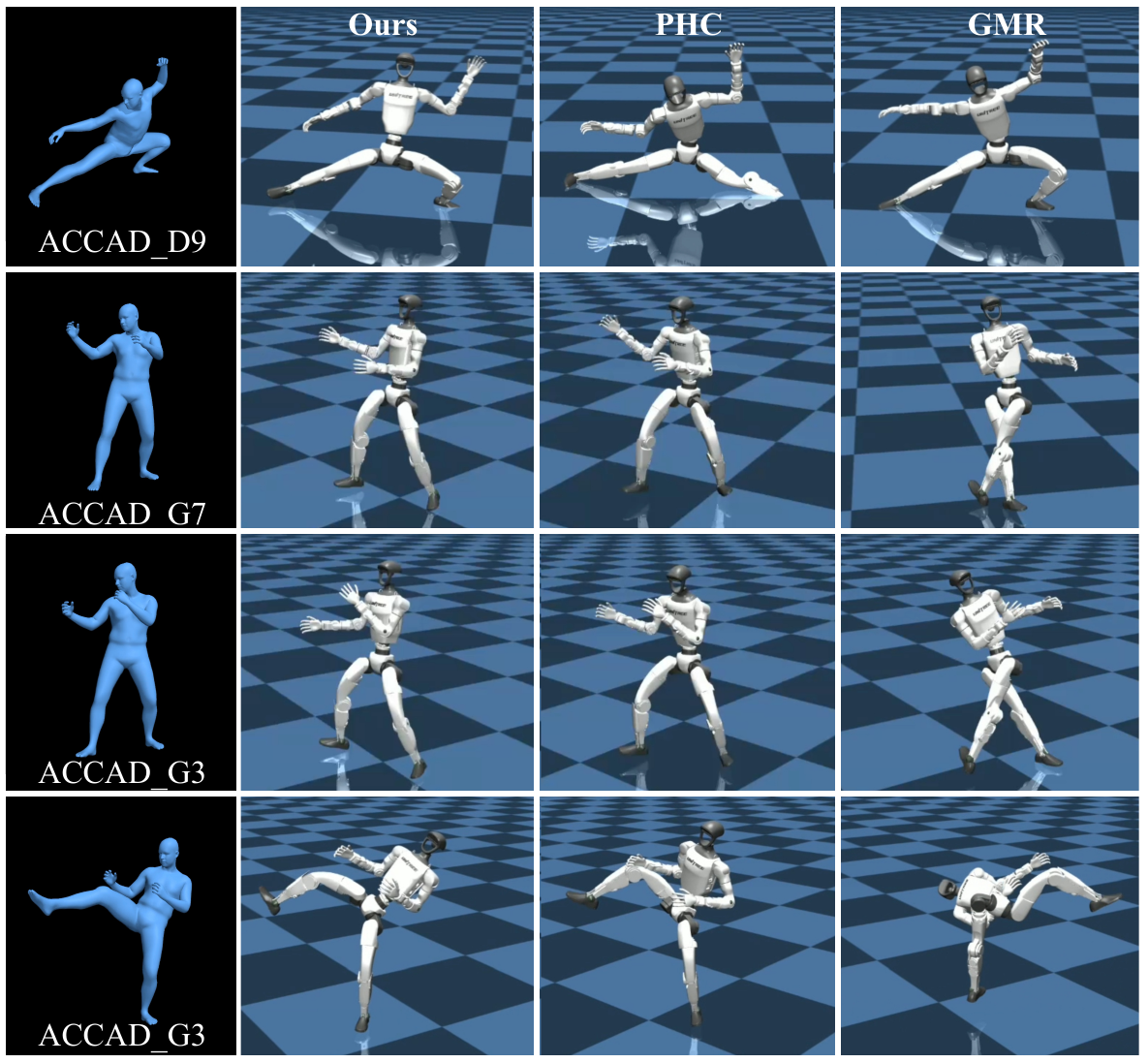}
    \caption{Qualitative comparisons between Human2Humanoid and the optimization-based baselines (PHC/GMR).}
    \label{fig:kongfu}
\end{figure}

Qualitative comparisons are also provided in Fig.~\ref{fig:kongfu}.
Compared with the optimization-based baselines (PHC/GMR), Human2Humanoid produces visually more stable contacts with fewer penetration artifacts under the large embodiment mismatches.

\subsection{Ablation Study}
\label{sec:ablation_qual}

To identify the contribution of each component, a controlled ablation study is conducted under the same human-to-Unitree-G1 retargeting setting as Table~\ref{tab:quantitative}. A separate set of evaluation motions is selected for the ablation analysis.
All ablation models share the same architecture, training schedule and evaluation protocol as the full model.
Specifically, the two considered ablation cases are:
(i) w/o $\mathcal{L}_{EE}$, which removes the morphology-invariant end-effector consistency loss while keeping the others unchanged,
and (ii) w/o $\mathcal{L}_{con}$, $\mathcal{L}_{hei}$, which removes the physics-aware contact and height constraints.
The same metric suite (SR, TE, FS and GP) is used to show how each component affects controllability and feasibility. The results are summarized in Table~\ref{tab:ablation}. Removing $\mathcal{L}_{EE}$ decreases the tracking success rate and increases tracking error, indicating that morphology-invariant end-effector consistency is important for preserving trackable motion semantics.
Removing the contact and height constraints increases both FS and GP and slightly worsens TE, indicating that the physics-aware objectives mainly improve contact-related feasibility, especially ground-penetration reduction.

\begin{table}[t]
\centering
\small
% \caption{Ablation study results. Values are mean $\pm$ std over 3 runs ($\downarrow$: lower is better, $\uparrow$: higher is better).}
\caption{Ablation study results ($\downarrow$: lower is better, $\uparrow$: higher is better).}
\label{tab:ablation}
\begin{tabular}{lcccc}
\hline
\textbf{Variant} 
& \textbf{SR (\%)} $\uparrow$
& \textbf{TE} $\downarrow$
& \textbf{FS (\%)} $\downarrow$ 
& \textbf{GP (cm)} $\downarrow$  \\
\hline
Full & 92.8 & 0.099 & 6.89 & 0.264 \\
w/o $\mathcal{L}_{EE}$ & 85.7 & 0.104 & 6.61 & 0.271 \\
w/o $\mathcal{L}_{con},\mathcal{L}_{hei}$ & 85.7 & 0.106 & 7.02 & 0.326 \\
\hline
\end{tabular}
\end{table}

%% file: tex/5-conclusion.tex
This paper presents \textbf{Human2Humanoid}, an unpaired motion retargeting framework for heterogeneous humanoid robots, achieving a more robust balance between semantic fidelity and physical executability. To address the pronounced human-robot discrepancies in skeletal topology and scale, as well as the scarcity of high-quality paired motion data, we build upon a CycleGAN-style unpaired learning architecture and incorporate skeleton-aware graph convolutional modeling. We further introduce a morphology-invariant end-effector consistency loss that uses the T-pose as reference and aligns cross-embodiment end-effector trajectories via scale normalization, thereby preserving motion semantics more reliably under large embodiment mismatches. Meanwhile, physical conditions such as contact consistency and joint limits are explicitly integrated into the training objective, which helps suppress deployment-critical artifacts including foot skating, base floating and ground penetration during generation. Experimental results show that Human2Humanoid enables high-quality retargeting without relying on paired data, and yields consistent improvements on both semantic and physical metrics. The ablation studies further validate the contribution of each key component. Future work will focus on generalizing to a broader range of non-humanoid robots with substantial topological differences, and on end-to-end coupling with downstream whole-body control policies to further improve real-robot robustness in complex interaction scenarios.

%% file: references.bib
@article{deepmimic,
   title={DeepMimic: example-guided deep reinforcement learning of physics-based character skills},
   volume={37},
   ISSN={1557-7368},
   url={http://dx.doi.org/10.1145/3197517.3201311},
   DOI={10.1145/3197517.3201311},
   number={4},
   journal={ACM Transactions on Graphics},
   publisher={Association for Computing Machinery (ACM)},
   author={Peng, Xue Bin and Abbeel, Pieter and Levine, Sergey and van de Panne, Michiel},
   year={2018},
   month=jul, pages={1–14} }

@article{chen2025gmt,
title={GMT: General Motion Tracking for Humanoid Whole-Body Control},
author={Chen, Zixuan and Ji, Mazeyu and Cheng, Xuxin and Peng, Xuanbin and Peng, Xue Bin and Wang, Xiaolong},
journal={arXiv:2506.14770},
year={2025}
}

@inproceedings{1998Retargetting,
author = {Gleicher, Michael},
title = {Retargetting motion to new characters},
year = {1998},
isbn = {0897919998},
publisher = {Association for Computing Machinery},
address = {New York, NY, USA},
url = {https://doi.org/10.1145/280814.280820},
doi = {10.1145/280814.280820},
booktitle = {Proceedings of the 25th Annual Conference on Computer Graphics and Interactive Techniques},
pages = {33–42},
numpages = {10},
series = {SIGGRAPH '98}
}

@article{2020Skeleton-aware,
author = {Aberman, Kfir and Li, Peizhuo and Lischinski, Dani and Sorkine-Hornung, Olga and Cohen-Or, Daniel and Chen, Baoquan},
title = {Skeleton-aware networks for deep motion retargeting},
year = {2020},
issue_date = {August 2020},
publisher = {Association for Computing Machinery},
address = {New York, NY, USA},
volume = {39},
number = {4},
issn = {0730-0301},
url = {https://doi.org/10.1145/3386569.3392462},
doi = {10.1145/3386569.3392462},
journal = {ACM Trans. Graph.},
month = aug,
articleno = {62},
numpages = {14}
}

@misc{choi2021selfsupervis,
      title={Self-Supervised Motion Retargeting with Safety Guarantee}, 
      author={Sungjoon Choi and Min Jae Song and Hyemin Ahn and Joohyung Kim},
      year={2021},
      eprint={2103.06447},
      archivePrefix={arXiv},
      primaryClass={cs.RO},
      url={https://arxiv.org/abs/2103.06447}, 
}

@article{Human–RobotInteraction,
    author = {Goodrich, Michael A. and Schultz, Alan C.},
    title = {Human–Robot Interaction: A Survey},
    journal = {Foundations and Trends in Human-Computer Interaction},
    volume = {1},
    number = {3},
    pages = {203-275},
    year = {2008},
    month = {01},
    issn = {1551-3955},
    doi = {10.1561/1100000005},
    url = {https://doi.org/10.1561/1100000005},
    eprint = {https://www.emerald.com/fthci/article-pdf/1/3/203/10914810/1100000005en.pdf},
}

@article{ayusawa2017motion,
  title={Motion retargeting for humanoid robots based on simultaneous morphing parameter identification and motion optimization},
  author={Ayusawa, Ko and Yoshida, Eiichi},
  journal={IEEE Transactions on Robotics},
  volume={33},
  number={6},
  pages={1343--1357},
  year={2017},
  publisher={IEEE}
}

@misc{darvish2019,
      title={Whole-Body Geometric Retargeting for Humanoid Robots}, 
      author={Kourosh Darvish and Yeshasvi Tirupachuri and Giulio Romualdi and Lorenzo Rapetti and Diego Ferigo and Francisco Javier Andrade Chavez and Daniele Pucci},
      year={2019},
      eprint={1909.10080},
      archivePrefix={arXiv},
      primaryClass={cs.RO},
      url={https://arxiv.org/abs/1909.10080}, 
}

@misc{zhang2023,
      title={Skinned Motion Retargeting with Residual Perception of Motion Semantics and Geometry}, 
      author={Jiaxu Zhang and Junwu Weng and Di Kang and Fang Zhao and Shaoli Huang and Xuefei Zhe and Linchao Bao and Ying Shan and Jue Wang and Zhigang Tu},
      year={2023},
      eprint={2303.08658},
      archivePrefix={arXiv},
      primaryClass={cs.CV},
      url={https://arxiv.org/abs/2303.08658}, 
}

@article{HuLei2024Pose,
  title={Pose-Aware Attention Network for Flexible Motion Retargeting by Body Part},
  author={HuLei and ZhangZihao and ZhongChongyang and JiangBoyuan and XiaShihong},
  journal={IEEE Transactions on Visualization and Computer Graphics},
  year={2024},
}

@article{2018Robust,
  title={Robust Real-Time Whole-Body Motion Retargeting from Human to Humanoid},
  author={ Penco, Luigi  and  Clement, B.  and  Moduano, V.  and  Hoffman, Enrico Mingo  and  Ivaldi, Serena },
  journal={IEEE},
  year={2018},
}

@article{2017Unpaired,
  title={Unpaired Image-to-Image Translation using Cycle-Consistent Adversarial Networks},
  author={ Zhu, Jun Yan  and  Park, Taesung  and  Isola, Phillip  and  Efros, Alexei A. },
  journal={arXiv e-prints},
  year={2017},
}

@misc{zhao2023posetomotion,
      title={Pose-to-Motion: Cross-Domain Motion Retargeting with Pose Prior}, 
      author={Qingqing Zhao and Peizhuo Li and Wang Yifan and Olga Sorkine-Hornung and Gordon Wetzstein},
      year={2023},
      eprint={2310.20249},
      archivePrefix={arXiv},
      primaryClass={cs.CV},
      url={https://arxiv.org/abs/2310.20249}, 
}

@misc{villegas2018neural,
      title={Neural Kinematic Networks for Unsupervised Motion Retargetting}, 
      author={Ruben Villegas and Jimei Yang and Duygu Ceylan and Honglak Lee},
      year={2018},
      eprint={1804.05653},
      archivePrefix={arXiv},
      primaryClass={cs.CV},
      url={https://arxiv.org/abs/1804.05653}, 
}

@misc{gmr,
      title={Retargeting Matters: General Motion Retargeting for Humanoid Motion Tracking}, 
      author={Joao Pedro Araujo and Yanjie Ze and Pei Xu and Jiajun Wu and C. Karen Liu},
      year={2025},
      eprint={2510.02252},
      archivePrefix={arXiv},
      primaryClass={cs.RO},
      url={https://arxiv.org/abs/2510.02252}, 
}

@misc{mao2017,
      title={Least Squares Generative Adversarial Networks}, 
      author={Xudong Mao and Qing Li and Haoran Xie and Raymond Y. K. Lau and Zhen Wang and Stephen Paul Smolley},
      year={2017},
      eprint={1611.04076},
      archivePrefix={arXiv},
      primaryClass={cs.CV},
      url={https://arxiv.org/abs/1611.04076}, 
}

@article{2017Motion,
  title={Motion Retargeting for Humanoid Robots Based on Simultaneous Morphing Parameter Identification and Motion Optimization},
  author={ Ayusawa, Ko  and  Yoshida, Eiichi },
  journal={Robotics, IEEE Trans. on (T-RO)},
  volume={33},
  number={6},
  pages={15},
  year={2017},
}

@misc{G-DReaM,
      title={G-DReaM: Graph-conditioned Diffusion Retargeting across Multiple Embodiments}, 
      author={Zhefeng Cao and Ben Liu and Sen Li and Wei Zhang and Hua Chen},
      year={2025},
      eprint={2505.20857},
      archivePrefix={arXiv},
      primaryClass={cs.RO},
      url={https://arxiv.org/abs/2505.20857}, 
}

@article{2019PMnet,
  title={PMnet: Learning of Disentangled Pose and Movement for Unsupervised Motion Retargeting},
  author={ Lim, Jongin  and  Chang, H.  and  Choi, J. },
  journal={British Machine Vision Association, BMVA},
  year={2019},
}

@misc{cheynel2025reconformrealtimecontactaware,
      title={ReConForM : Real-time Contact-aware Motion Retargeting for more Diverse Character Morphologies}, 
      author={Théo Cheynel and Thomas Rossi and Baptiste Bellot-Gurlet and Damien Rohmer and Marie-Paule Cani},
      year={2025},
      eprint={2502.21207},
      archivePrefix={arXiv},
      primaryClass={cs.GR},
      url={https://arxiv.org/abs/2502.21207}, 
}

@misc{kim2025moreflowmotionretargetinglearning,
      title={MoReFlow: Motion Retargeting Learning through Unsupervised Flow Matching}, 
      author={Wontaek Kim and Tianyu Li and Sehoon Ha},
      year={2025},
      eprint={2509.25600},
      archivePrefix={arXiv},
      primaryClass={cs.GR},
      url={https://arxiv.org/abs/2509.25600}, 
}

@misc{Motion-X,
      title={Motion-X: A Large-scale 3D Expressive Whole-body Human Motion Dataset}, 
      author={Jing Lin and Ailing Zeng and Shunlin Lu and Yuanhao Cai and Ruimao Zhang and Haoqian Wang and Lei Zhang},
      year={2024},
      eprint={2307.00818},
      archivePrefix={arXiv},
      primaryClass={cs.CV},
      url={https://arxiv.org/abs/2307.00818}, 
}

@misc{PHUMA,
      title={PHUMA: Physically-Grounded Humanoid Locomotion Dataset}, 
      author={Kyungmin Lee and Sibeen Kim and Minho Park and Hyunseung Kim and Dongyoon Hwang and Hojoon Lee and Jaegul Choo},
      year={2025},
      eprint={2510.26236},
      archivePrefix={arXiv},
      primaryClass={cs.RO},
      url={https://arxiv.org/abs/2510.26236}, 
}

@inproceedings{phc,
    author={Zhengyi Luo and Jinkun Cao and Alexander W. Winkler and Kris Kitani and Weipeng Xu},
    title={Perpetual Humanoid Control for Real-time Simulated Avatars},
    booktitle={International Conference on Computer Vision (ICCV)},
    year={2023}
}

@misc{nmr,
      title={Make Tracking Easy: Neural Motion Retargeting for Humanoid Whole-body Control}, 
      author={Qingrui Zhao and Kaiyue Yang and Xiyu Wang and Shiqi Zhao and Yi Lu and Xinfang Zhang and Wei Yin and Qiu Shen and Xiao-Xiao Long and Xun Cao},
      year={2026},
      eprint={2603.22201},
      archivePrefix={arXiv},
      primaryClass={cs.RO},
      url={https://arxiv.org/abs/2603.22201}, 
}

@article{Globalinversekinematics,
author = {Dai, Hongkai and Izatt, Gregory and Tedrake, Russ},
title = {Global inverse kinematics via mixed-integer convex optimization},
year = {2019},
issue_date = {Oct 2019},
publisher = {Sage Publications, Inc.},
address = {USA},
volume = {38},
number = {12–13},
issn = {0278-3649},
url = {https://doi.org/10.1177/0278364919846512},
doi = {10.1177/0278364919846512},
journal = {Int. J. Rob. Res.},
month = oct,
pages = {1420–1441},
numpages = {22}
}
